\documentclass[conference]{IEEEtran}
\IEEEoverridecommandlockouts
\usepackage{cite}
\usepackage{amsmath,amssymb,amsfonts}
\usepackage{algorithmic}
\usepackage{graphicx} 
\usepackage{textcomp}
\usepackage{xcolor}
\def\BibTeX{{\rm B\kern-.05em{\sc i\kern-.025em b}\kern-.08em
    T\kern-.1667em\lower.7ex\hbox{E}\kern-.125emX}}

\usepackage{caption} 
\usepackage{booktabs}
\usepackage{colortbl}

\usepackage{array,tabularx}

\usepackage{subcaption}
\usepackage{float}
\usepackage[english]{babel}
\usepackage{blindtext}
\usepackage[utf8x]{inputenc}

\usepackage[super]{nth} 
\usepackage[inline]{enumitem}

\usepackage[hyphens]{url}
\usepackage[hidelinks]{hyperref}
\hypersetup{breaklinks=true}
\graphicspath{{graphics/}}

\usepackage[font=small,labelfont=normalfont]{caption}
\usepackage[ruled,linesnumbered,vlined]{algorithm2e}
\usepackage{xcolor}

\SetCommentSty{mycommfont}

\renewcommand\IEEEkeywordsname{Keywords}

\begin{document}

\title{A new hashing based nearest neighbors selection technique for big datasets}

\author{\IEEEauthorblockN{Jude~Tchaye-Kondi}
\IEEEauthorblockA{\textit{School of Computer Science and} \\
	\textit{Technology} \\
\textit{Beijing Institute of Technology}\\
Beijing, China \\
tchaye59@gmail.com}
\and
\IEEEauthorblockN{Yanlong~Zhai}
\IEEEauthorblockA{\textit{School of Computer Science and} \\
	\textit{Technology} \\
\textit{Beijing Institute of Technology}\\
Beijing, China \\
ylzhai@bit.edu.cn}
\and
\IEEEauthorblockN{Liehuang~Zhu}
\IEEEauthorblockA{\textit{School of Computer Science and} \\
	\textit{Technology} \\
\textit{Beijing Institute of Technology}\\
Beijing, China \\
liehuangz@bit.edu.cn;}
}

\maketitle

\begin{abstract}
KNN has the reputation to be the word simplest but efficient supervised learning algorithm used for either classification or regression. KNN prediction efficiency highly depends on the size of its training data but when this training data grows KNN suffers from slowness in making decisions since it needs to search nearest neighbors within the entire dataset at each decision making. This paper proposes a new technique that enables the selection of nearest neighbors directly in the neighborhood of a given observation. The proposed approach consists of dividing the data space into subcells of a virtual grid built on top of data space. The mapping between the data points and subcells is performed using hashing. When it comes to select the nearest neighbors of a given observation, we firstly identify the cell the observation belongs by using hashing, and then we look for nearest neighbors from that central cell and cells around it layer by layer. From our experiment performance analysis on publicly available datasets, our algorithm outperforms the original KNN in time efficiency with a prediction quality as good as that of KNN it also offers competitive performance with solutions like KDtree.
\end{abstract}


\section{Introduction}
\begin{figure}[H]
	\centering
	\includegraphics[width=0.9\linewidth]{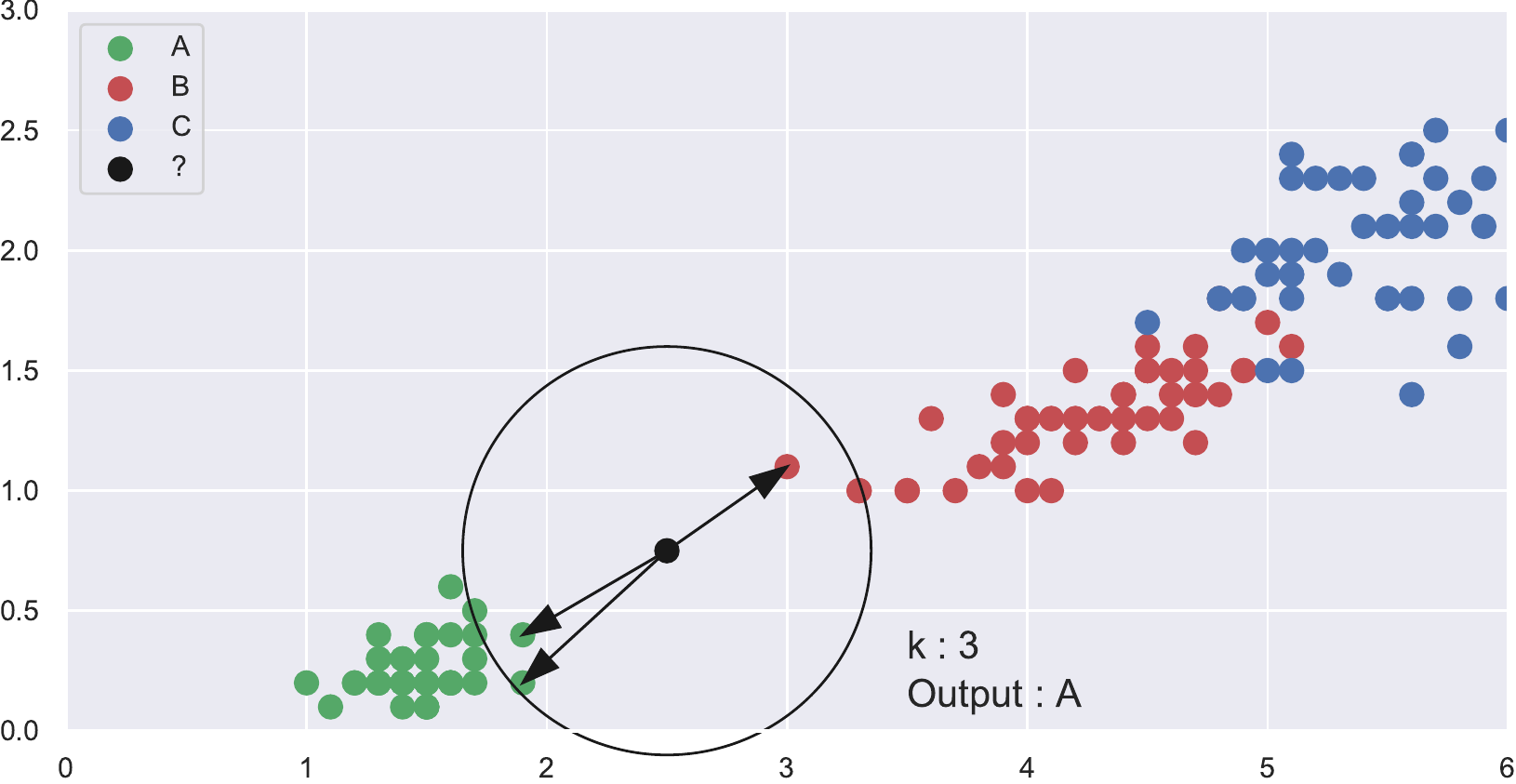}
	\caption{KNN example}
	\label{fig:knn_example}
\end{figure}

\IEEEPARstart{T}{he} K nearest neighbors or simply KNN is an algorithm that works on a very simple principle that is: tell me who are your neighbors and I will tell you who you are(Figure \ref{fig:knn_example}). So KNN does not need a statistical model to be able to make a prediction, it learns nothing from the training data and has to carry the full dataset during its decision making. For this reason, KNN is categorized as a Lazy Learning algorithm.
The K of KNN is not a parameter but a hyperparameter because unlike conventional parameters, it will not be learned automatically by the algorithm from the training data but it is up to us to optimize it, using the test data set. To predict an observation, which is not part of the training dataset, the algorithm will look for the K closest instances of the dataset to our observation. Then for these neighbors, the algorithm will be based on their output values to predict the output of the observation that we want to predict. Accordingly:
\begin{itemize}
	\item If KNN is used for the regression, it is the average or median of the K nearest observations outputs that will be used for the prediction
	\item If KNN is used for classification, it is the majority class among the K nearest observation classes that will be used for the prediction
\end{itemize}

From the simple classification example shown in Figure \ref{fig:knn_example} with k = 3, KNN will predict A for our input since class A is the majority element among the 3 nearest elements.
KNN is a supervised learning method with very good prediction accuracy, hence its wide use in several domains.
In medicine, researchers proposed a KNN based drug classification approach which is used to categories the different types of drug\cite{chduknnga}, it is also used for diagnosing diseases like Heart disease patients\cite{aknnidhdp}, cancer prediction, and detection\cite{medjahed2013breast, 7150592, jabbar2017prediction }. In image processing, KNN work efficiently in image classification\cite{amato2010knn}, face recognition\cite{ebrahimpour2007face}. The KNN algorithm is also present in cybersecurity where it is effectively used for credit card fraud detection\cite{ganji2012credit}, detection of intrusive attacks in a network system\cite{liao2002use}.
The strength of KNN is that in addition to being a very simple algorithm, it is very effective, but behind this effectiveness, hides two major weaknesses that are:
\begin{itemize}
	\item The big size of the model: As KNN does not use a statistical model, a KNN model needs to carry the whole training dataset.
	\item Prediction slowness: During prediction, KNN needs to browse the whole dataset for distance computations before selecting the nearest neighbors. This process has a time complexity of $(fn)log(fn)$ with $n$ the training dataset size and $f$ the number of features.
\end{itemize}
KNN is also very sensitive to noises(outliers) and highly dependent on the choice of its parameter k and the chosen distance metric. The slowness and model size weaknesses put lots of limits on the use of KNN with large training data.
With the fact that KNN runs in memory and that RAM is limited,  it will be difficult with this algorithm to maintain a big dataset model in memory. By the slowness that results, this algorithm is not appropriate for real-time applications or applications that have strict time limits requirements.
In this paper, we focused on the slowness problem of KNN during predictions with big training datasets. To improve the prediction time efficiency, we propose a new hashing based algorithm that we called GHN: Grid Hashing Neighborhood. GHN approach consists in splitting the data space into sub-cells by building a virtual grid on top of it. A mapping hash table between the data points and cells is then built at the learning phase and is then used during prediction to fast the nearest neighbor's selection.
During prediction, the nearest neighbors of a new observation are selected in two steps. Firstly, the central cell to which the new observation belongs is determined using a hash function, and secondly, we search for nearest neighbors from this central cell to its neighbor cells layer by layer.
Unlike with the native KNN where we have to go through all the training data points, GHN enables the selection of nearest neighbors directly in the neighborhood of our observation. Our performance analysis shows that our approach is faster in making predictions than the native KNN.

The rest of the paper is structured as follows.
Section \ref{sec:RelatedWork} reviews related work.
Section \ref{sec:Solution} the methodology of GHN.
Section \ref{sec:ExperimentalEvaluation} evaluates the performance of our implementation of GHN, the original KNN, and KDtree on some publicly available datasets.
Finally, the paper is concluded in Section \ref{sec:Conclusion}.

\section{Related Work}
\label{sec:RelatedWork}

The slowness of KNN predictions is not a new problem to the machine learning field, unlike humans who just by looking at the data representation in a 2D or 3D vector space can guess by intuition the closest neighbors of a data point, a computer requires more calculations for the same task. Three main groups zzemerge among the different techniques used to compute the nearest neighbors which are: data reduction approaches, hashing based approaches, and tree-based approaches.
The major portion of these solutions is composed of data reduction techniques which objective is to reduce the training data size, thereby reducing the number distance computations during prediction and the amount of required memory by the model. Reducing the training data size can be very effective for certain types of datasets that still can perform accurately with certain particular data points as the dataset. These techniques are less adopted despite the fact they are faster in prediction and improve memory usage. This is because these data reduction techniques usually do not offer the same quality of prediction accuracy as KNN.

In \cite{09067590200403835, 10.1117/12.873054,chaknnaftcotrobasop, anchaacmfndd} concave and convex hulls based techniques are proposed and used to reduce each class samples to their edge data points. In these techniques, only the edge points are used in the training datasets for classification.
Hart et Al. proposed the condensed nearest neighbor(CNN)\cite{hart1968condensed} which reduces its data by selecting prototypes U from the training data in a way that 1NN with U can classify the samples almost as precisely as 1NN does with the dataset. CNN works in 3 steps\cite{cknna}:
\begin{enumerate*}
	
	\item Scans all the elements of the training data X, looking for an element x whose nearest prototype from U has a label different from x.
	
	\item Remove x from X and add it to U.
	
	\item Repeat the operation until no other prototype is added to U. In the end use U instead of X to train the model.
	
\end{enumerate*}
Salvador et Al. introduce compressed kNN \cite{salvador2019compressed} which is a binary level data compression technique. The method proposes to compress observations into packets of a certain number of bits, in each packet a certain number of attributes are stored through binary level operations. This technique reduces the amount of RAM needed to maintain the training data in memory. An interesting feature of the compressed kNN method is that the information can be decompressed, observation by observation on-the-fly and in real-time, without the need to decompress all the dataset and carry out it into the memory. But unfortunately, compressed kNN still suffers from slowness and only works with categorical data.

During our research, we noticed that there is not enough proposed work based on using hashing techniques to compute nearest neighbors. Hashing is used to group similar data points in buckets. The most popular hashing based solution is the LSH(Locality Sensitive Hashing) family \cite{pan2011fast}\cite{zhang2013fast}\cite{bagui2019improving}\cite{wu2019fast}. LSH based solutions use random plane projections in the data space to divide that space into sub-regions. These sub-regions are then used as a bucket to build a hash function. Even if LSH based solutions improve prediction time, the strategy behind them is ineffective and does not guarantee to get the real nearest neighbors hence its low adoption. Gao et Al. try to face this problem by suggesting another family of hashing technique that is DHT \cite{gao2014dsh} which, unlike LHT family, can maintain relationships between the nearest neighbors.
Tree-based solutions are the most adopted in real-world problems when it comes to approximating the nearest neighbors. The most famous are KD tree\cite{sproull1991refinements}\cite{Zhou:2008:RKC:1409060.1409079} and Ball Tree\cite{omohundro1989five}\cite{liu2006new}. They are data structures that organize training data like a tree. When searching for the nearest neighbors, we navigate the tree from top to bottom hoping that the region we led in will contain the nearest neighbor. Just like LSH, these tree-based solutions can easily miss the real nearest neighbor and they are mostly recommended for low dimensional space since they don’t perform with multi-dimensions.
There is still no efficient solution for computing with precision the KNN which offers both a low computations cost and a low memory cost. The existing one suffers from drawbacks like degradation of the quality of prediction (accuracy) or the risk of having fake nearest neighbors. Our solution adopts a unique hashing based approach that allows us to directly select our neighbors around the observation during prediction by assuring good performances.

\section{METHODOLOGIES}
\label{sec:Solution}

\begin{figure*}[ht] 
	
	\centering
	
	\includegraphics[width=\linewidth]{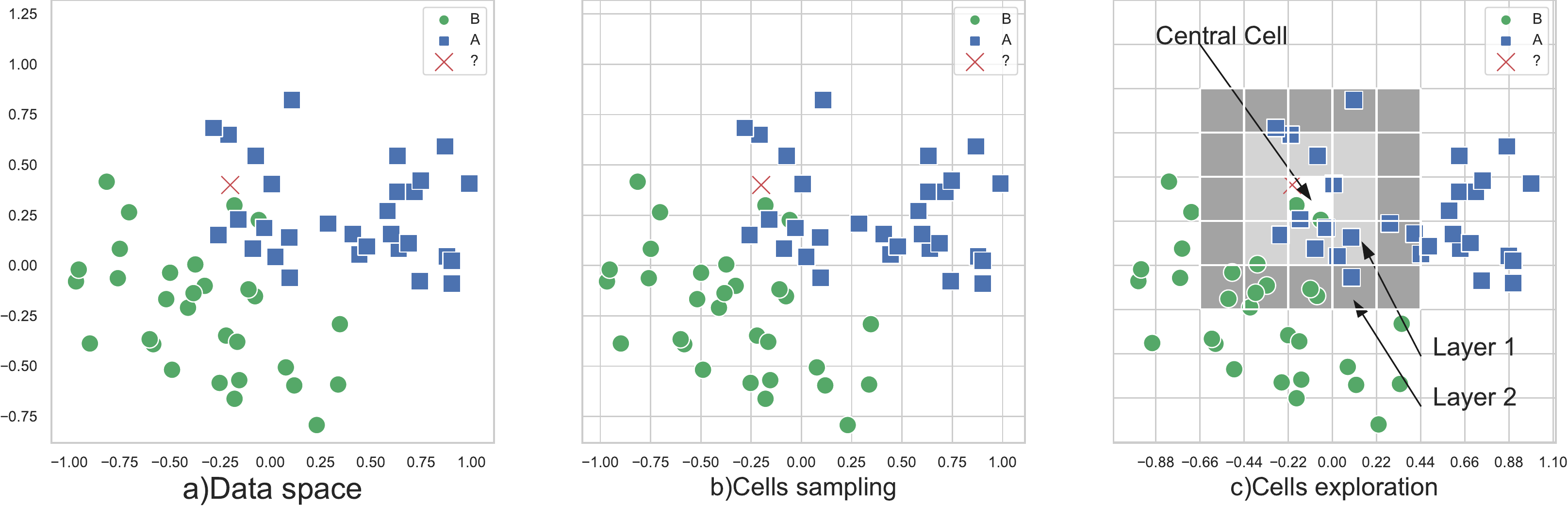}
	
	\caption{Different steps off GHN algorithm}
	
	\label{fig:gridknn}
	
\end{figure*}

Compared to the other machine learning algorithms KNN does not have a learning phase, the dataset does not undergo any transformation and is entirely maintained in memory. It is during predictions that KNN does all of its computations (distances, nearest neighbors’ selection). Unlike KNN, GHN has a learning phase before the prediction one. 
To illustrate the GHN algorithm we will take the simple case of a two-dimensional space, i.e. when the training data only have two features since it is much easier to visualize and understand (Figure \ref{fig:gridknn}.a). This can be generalized to multiple dimensions.
Figure \ref{fig:gridknn}.a contains the observations of two classes, the class A observations are represented by blue squares, that of class B by green circles and the new observation whose class is to be predicted is represented by a red cross. For this example, $k = 3$, so our goal is to find the 3 nearest data points to our new observation.
GHN algorithm consists mainly of two steps: 
\begin{enumerate}
	\item \textbf{Cells sampling:} this phase is performed during training. We subdivide the data space into identical sub-cells by building on top of it a virtual grid as illustrated in {Figure \ref{fig:gridknn}.b }. A Hash function is used to map training data points to their corresponding sub-cells.
	
	\item \textbf{Exploration:} this phase consists of selecting the nearest neighbors of a given input. As shown in {Figure \ref{fig:gridknn}.c}, GHN firstly determines the sub-cell to which the new observation with unknown output belongs by using the hash function and secondly searches nearest neighbors from data points in this central cell and cells in its neighborhood layers by layer.
	
\end{enumerate}

\subsection{Cell sampling}
\begin{figure}[H]
	
	\centering
	
	\includegraphics[width=\linewidth]{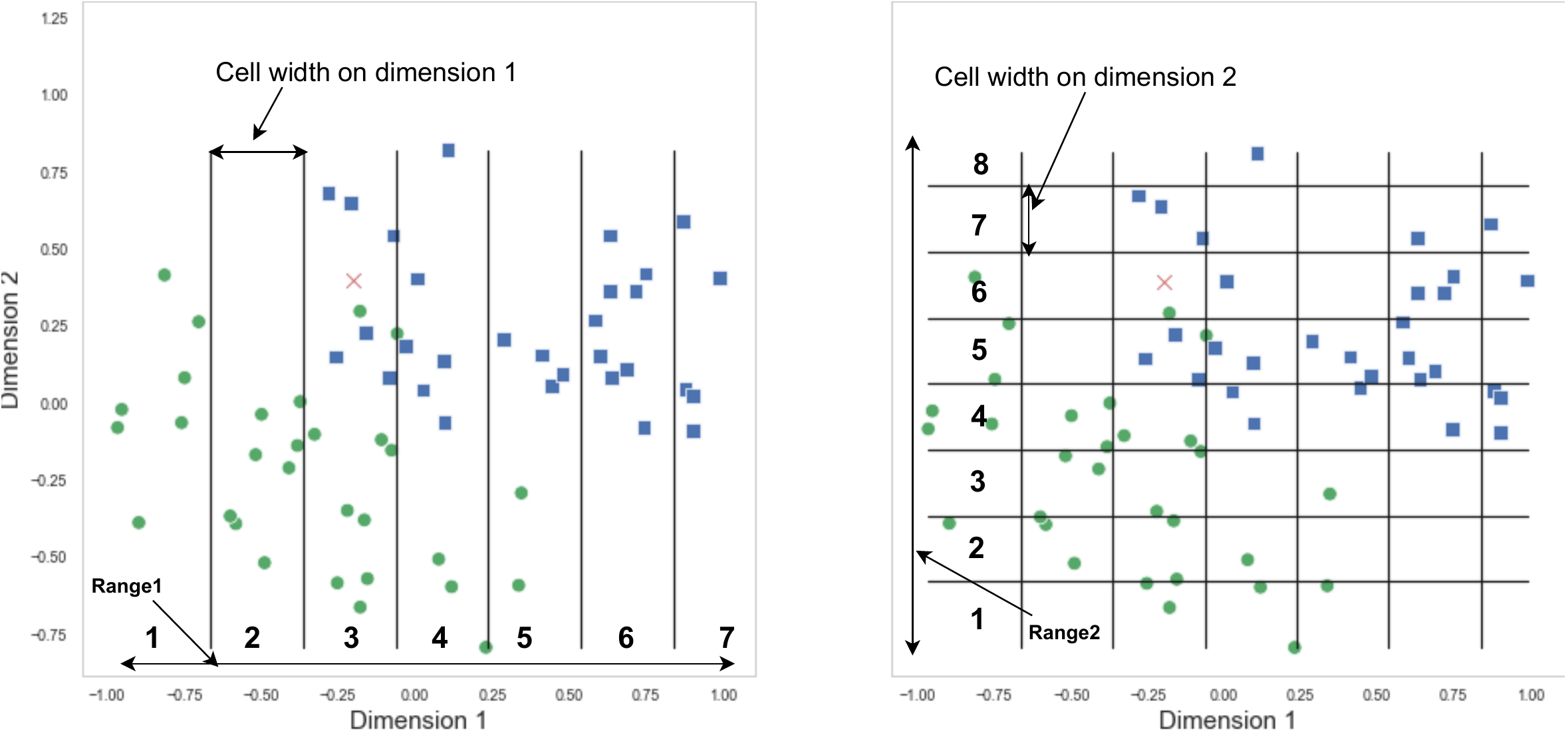}
	
	\caption{Cell measurements}
	
	\label{fig:cl_measurements}
	
\end{figure}

This is done during the training step of the model. As said above, during this phase, we build a virtual grid on top of the data space to split it into subcells that will contain the data point located in the area covered by the cell. These cells represent the buckets of our hash table, they are identical but not necessarily equilateral. 
To build the virtual grid, we need to determine the cell measurements on each dimension. 
To do this, we divide values range covered by the training data points on each dimension in a maximum possible splits in a way that each split ends up with at least one data point. The cell measurements on the corresponding dimension are the split width.
This process is illustrated with{ Figure \ref{fig:cl_measurements}}, in this figure, the first dimension can have a maximum of 7 splits and 8 splits on the second dimension so the cell measurements will be (range1/7, range2/8)  on the first and second dimensions respectively.
This way of determining the virtual grid's cell measurements allows ensuring an optimal distribution of the data points in cells and also ease the lookup of nearest neighbors during exploration. Once the measurements of the cell on each dimension are determined, a hash table mapping data points to their corresponding cells are built by using Equation \ref{equ:cell_id}. Each cell in the virtual grid has a unique identifier and this identifier can be computed from a data point or any new observation by using the Equation \ref{equ:cell_id}.
Data points that belong to the same cell have the same cell\_id. GHN hash table does not keep any information about empty cells for memory efficiency. The cell sampling process is simplified with Algorithm \ref{code:sampling_algo}.

\begin{algorithm}[!ht]
	\SetAlgoLined
	$grid$ : A hash table where the key is cellId\;
	$a$ : cell measurements\;
	$data$ : training data\;
	
	\tcc{Model Training : Cells sampling}
	
	\For {$point$ in $data$}{
		cellId = $point // a$\;
		\If{$cellId$ not grid}{
			\tcc{Initialize the cell}
			grid[cellId] = [];
		}
		grid[cellId].append(point)\;
	}
	
	\caption{Cell sampling Algorithm}
	\label{code:sampling_algo}
\end{algorithm}

\begin{equation}
	cell\_id = P // a
	\label{equ:cell_id}
\end{equation}
\begin{itemize}
	\item $P$: The data point.
	\item $//$: Integer division.
	\item $a$: Cell Measurements
\end{itemize}

\subsection{Exploration}

The actual neighbors' selection is done during this exploration step. The idea is to start from the cell to which our new observation belongs, knowing that cell id defines by Equation \ref{equ:cell_id}, we can compute the ids of cells in its neighborhood directly. The two steps of the exploration are:  
\begin{enumerate}
	\item Get the central cell, the one containing the new observation using Equation \ref{equ:cell_id}.
	\item Fetch the data points from the central cell and its neighbors' layer by layer as shown by the example of Figure \ref{fig:gridknn}.c.
\end{enumerate} 
The central cell id is computed by applying equation \ref{equ:cell_id} on the new observation and after, the data is recovered from the hash table.
In our implementation, we use the breadth-first search\cite{elmasry2015space} (BFS) searching techniques to compute ids of cells in the neighborhood of the central cell on each layer and visit them. 
When a cell is visited its data points are collected in a buffer. This buffer is a heap of size k which only keeps the k nearest element by using heap sort mechanism \cite{schaffer1993analysis }. 
The exploration stop when a layer is visited and there is no update among the buffer elements. 
In Figure \ref{fig:gridknn}.c where K = 3: 
\begin{itemize}
	\item Exploration of the central cell: The central cell only contains 1 data point. We add this observation to our buffer and since the buffer size which is one is less than 3, we explore the cells on the first layer.
	\item Exploration of the First layer: The first layer is composed of 8 cells. After visiting these 8 cells, from the data point collected from the visited cells, the buffer will only keep the 3 nearest data points to our observation. The next step is to visit the second layer.
	\item Exploration of the second layer: The second layer is composed of 16 cells. Visiting these cells doesn’t provide any update among the data point in the buffer so the exploration stops there.
\end{itemize}
At the end of exploration, the buffer will remain with exactly k element which are the nearest neighbors to our observation. It is important to note that in some extremely rare cases, the exploration may miss the real nearest neighbors this is due to the fact the virtual grid cells don’t have the same measurements on each dimension.
The whole exploration process is illustrated by algorithm \ref{code:grid_knn}.

\begin{algorithm}[!ht]
	\SetAlgoLined
	$grid$ : The hash table \;
	$a$ : cell measurements\;
	$data$ : training data\;
	
	\tcc{Select K Nearest: Cells exploration}
	
	$central\_cellId$ = $new\_input // a$\;
	
	$queue$ = [$central\_cellId$]\;
	$buffer$ = []\;
	
	\While {queue is not empty}{
		cellId = queue.dequeue()
		
		\If{$cellId$ not $grid$}{
			buffer.addAll(grid[cellId])
		}
		Add neighbors’ cells to the queue\;
		\For {$next\_cell$ in $neighborsCells$}{
			queue.enqueue(next\_cell)
		}	
	}
	
	\Return buffer\;

	\caption{Exploration Algorithm}
	\label{code:grid_knn}
\end{algorithm}

\subsection{Performances comparison with KNN}

The complexity of GHN mainly depends on the number of features and the depth of the exploration (visited layers) which is strongly influenced by data distribution. GHN compared with existing solutions can achieve record performance with big training datasets. The more the data grow the faster is GHN  since it directly searches for neighbors in the neighborhood of the observation. Its performance is almost constant (O(1)) when the input is located in a densely populated region of the data space, if not GHN will require a little more effort and explore more.
The time taken by GHN to make a prediction is the time taken by its exploration phases added to the time required to process the buffer that contains the k nearest elements:

\begin{eqnarray}
	Time\ Complexity = Exploration\ +  Buffer\ processing\nonumber
\end{eqnarray}
\begin{itemize}
	
	\item Exploration time: Depend on the depth of the exploration and the number of data points processed from visited cells. With the depth of the exploration, we can define the total number of visited cells with {Equation \ref{equ:total_cell_number}}.
	
	\item Buffer processing time: it takes $O(k)$ time to process the buffer because, after the exploration phase, the buffer only remains with k elements.
	
\end{itemize}
During the exploration the number of cells on each layer is given by the following Equation \ref{equ:layer_cell_number} and proved in Appendix \ref{AppendixA}.
\begin{equation}
	n = (2l+1)^{d} - (2l-1)^{d}
	\label{equ:layer_cell_number}
\end{equation}Where:
\begin{itemize}
	\item $l$: the number of visited layers.
	\item $d$: the number of dimensions or features.
\end{itemize}
Thus, if the exploration stop after $l$ layer, the total number of cells from the central cell this layer is expressed by the following Equation \ref{equ:total_cell_number} . We also proved it in Appendix \ref{AppendixB}:

\begin{equation}
	S_{cells}(l) = (2l+1)^d - 1
	\label{equ:total_cell_number}
\end{equation}
The number of features $d$ is fixed and does not change during the use of the model, then only the exploration depth $l$ is influencing $S_{cells}(l)$.
We cannot extrapolate the different values of the exploration depth and the number of data points processed during the exploration phase since they depend on the data distribution. This makes it difficult to directly compare GHN with the KNN, nevertheless, we will rather look at the best-case comparison of them. The best case that GHN allows is when the exploration is done in a very dense area and stop after exploring a single layer ($depth = l = 1$) and collecting k data point.
For this best-case scenario, the number of cells visited is obtain by setting $l = 1$ in Equation \ref{equ:total_cell_number}:
\begin{eqnarray}
	S_{1} = 3^{d}-1  \nonumber
\end{eqnarray}
Then the best-case complexity of is:
\begin{eqnarray}
	O(3^d -1)+O(k) ≃O(1)+O(1) ≃O(1)  \nonumber
\end{eqnarray}
\begin{itemize}
	\item $O(3^d -1)$ is the cell exploration complexity. $d$ can be neglected since it is a constant: $O(3^d -1) ≃O(1)$
	\item $O(k)$ is the complexity for processing  buffer. $k$ also is a constant and can be neglected.
\end{itemize}

The proposed approach can reach a best-case complexity of almost $O(1)$ which is far better than the original KNN and which until now has not been possible with all the solutions proposed so far. With original KNN, the best and worst-case time complexity is $O(ndlog(dn)))$. GHN and KNN have both memory complexity of $O (dn)$. Even if GHN uses slightly more memory than KNN to store its hash table, the number of cells never exceeds n (the size of the training dataset). GHN is the only proposed solution whose predictions time performance is not negatively influenced by the growth in the training data size. With existing solutions, the more the dataset size increases the slower they are but with our approach, the more the learning data increases, the better.

\subsection{Discussion}

\begin{figure}[H]
	\centering
	\includegraphics[width=\linewidth]{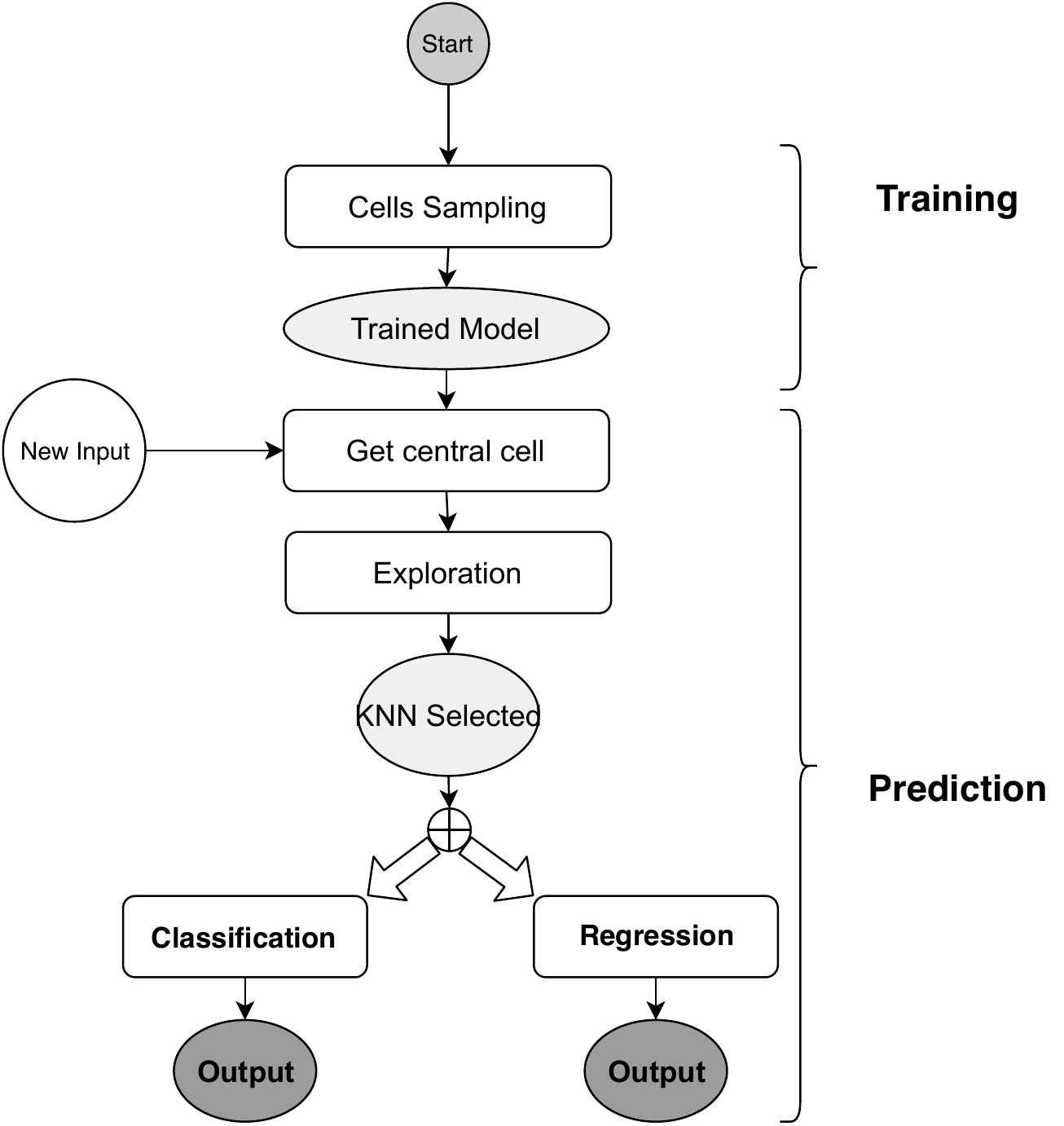}
	\caption{GHN Flow}
	\label{fig:gridknn_draft}
\end{figure}

The proposed solution can be used for classifications as well as for regressions tasks as it is only intended to improve the selection of nearest neighbors. In the draft of Figure \ref{fig:gridknn_draft} we can see that once the nearest neighbors selected, they are used as input of a classification model that is responsible for predicting the majority class by a vote or a regression model that will predict the average or the median of the k selected samples. Features in the training dataset take their values from completely different scales of magnitude. It is recommended to rescale the training data to put features on the same magnitude. This is performed using data scaling techniques like Standardization, Mean Normalization, Unit Vector, etc.
It is important to note that GHN is not efficient for all types of datasets as it assumes that all data points converge in the same area of the data space. When the data points are far away from each other or when our observation is very far from the region where converge the data points, GHN's performance degrades. As KNN, GHN is also subject to the curse of dimensionality \cite{hinrichs2010curse}. This happens when the number of states exponentially increases for a tiny increase in the number of dimensions or parameters due to a combinatory explosion. This is noticed in Equation \ref{equ:total_cell_number} where each time the number of layers increases, the number of cells to be visited is an exponential function of the dimensions.

\section{Experiments and Results}
\label{sec:ExperimentalEvaluation}

We evaluate the GHN performance against the original KNN and the popular KDTree. Our experiment target two main aspects that are the prediction time with which we evaluate the time efficiency and the prediction quality by using an accuracy score metric.
Our tests are performed on 5 real dataset that we grab from different datasets repository\cite{Dua:2019, lecun2010mnist, winequalityred, RussianDemography, PulsarStar, xiao2017/online}, these datasets are presented in {Table \ref{tab:datasets}}.  Each dataset is used for classification tasks only in order to facilitate comparisons. We scale the datasets and reduce their dimensions by using Principal component analysis(PCA). Our experiment is implemented in python 3.6 and the running environment is an Ubuntu laptop of processor Intel i7.
\begin{table}[H]
	
	\centering
	
	\arrayrulecolor[rgb]{0.749,0.749,0.749}
	
	\resizebox{\linewidth}{!}{%
		
		\begin{tabular}{ll} 
			
			\arrayrulecolor{black}\hline
			
			\multicolumn{1}{l}{Datasets}                                              & Description                                                                                                                                                                                      \\ 
			
			\hline
			
			Fashion MNIST\cite{xiao2017/online}                                             & \begin{tabular}[c]{@{}l@{}}Fashion-MNIST is a dataset of Zalando's article images. \\Each example is a 28x28 grayscale image, associated \\with a label from 10 classes.\end{tabular}            \\ 
			
			\arrayrulecolor[rgb]{0.749,0.749,0.749}\hline
			
			MNIST\cite{lecun2010mnist}                                                      & \begin{tabular}[c]{@{}l@{}}MNIST database of handwritten digits has a training set \\of 60,000 examples. The digits have been size-normalized\\and centered in a fixed-size image.\end{tabular}  \\ 
			
			\hline
			
			\begin{tabular}[c]{@{}l@{}}Pulsar Star\cite{PulsarStar,Dua:2019}\end{tabular} & \begin{tabular}[c]{@{}l@{}}Describes a sample of pulsar candidates collected \\during the High Time Resolution Universe Survey.\end{tabular}                                                     \\ 
			
			\hline
			
			Wine Quality\cite{winequalityred}                                               & Data tells various chemical combination of red wine                                                                                                                                              \\ 
			
			\hline
			
			Russian Demography\cite{RussianDemography}                                      & \begin{tabular}[c]{@{}l@{}}Russian Demography (1990-2017) Dataset. \\It contains demographic features like natural population \\growth, birth rate, population, etc.\end{tabular}                \\
			
			\arrayrulecolor{black}\hline
			
		\end{tabular}
		
	}
	
	\label{tab:datasets}
	
	\caption{Datasets}
	
\end{table}

\begin{figure}[H]
	
	\centering
	
	\includegraphics[width=0.9\linewidth]{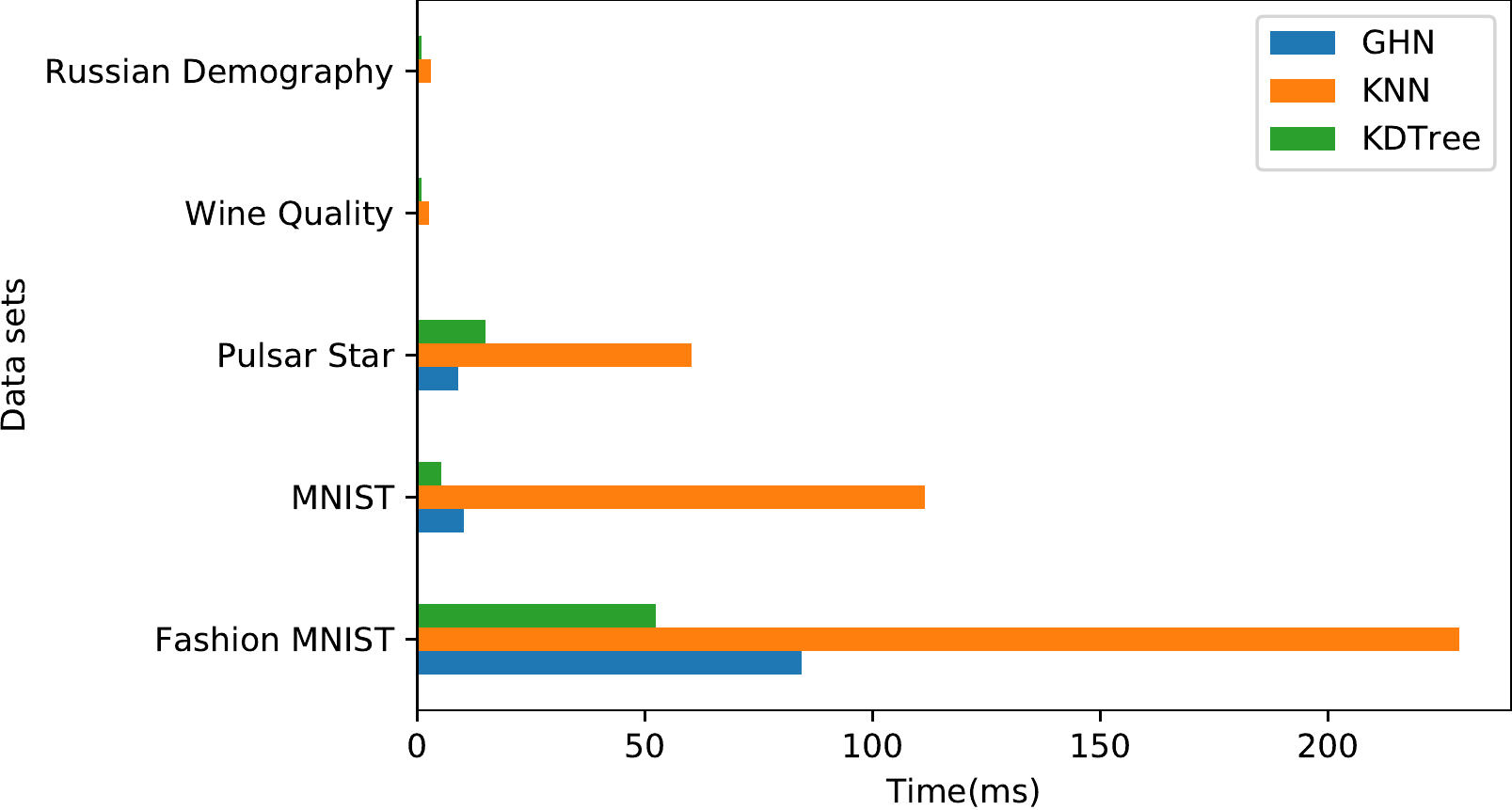}
	
	\caption{Comparison of prediction times on various datasets}
	
	\label{fig:times}
	
\end{figure}

We have collected in Table \ref{tab:times} the time taken by each model to evaluate the test data for each dataset.
\begin{table}[H]
	\centering
	\resizebox{0.7\linewidth}{!}{%
		\begin{tabular}{|l|l|l|l|} 
			\hline
			Datasets & GHN & KDTree & KNN \\ 
			\hline
			Fashion MNIST & 84.29 & 52.40 & 228.77 \\
			MNIST & 10.00~ & 5.22 & 111.48 \\
			Pulsar Star & 9.09 & 15.00 & 60.20 \\
			Wine Quality & 0.27 & 0.87 & 2.46 \\
			Russian Demography & 0.19 & 0.90 & 2.92 \\
			\hline
		\end{tabular}
	}
	\caption{Prediction Times(ms) on various datasets}
	\label{tab:times}
\end{table}
For each dataset, 80\% is used for training and the remaining 20\% for testing. We can see in Figure \ref{fig:times} that GHN and KDtree are much faster than the original KNN on the 5 datasets.
We also notice that KDT is slightly faster than GHN on image  data type, this is the effect of the curse of dimensionality which GHN faces during the exploration since a flatten image end up with a high dimensional vector. To face this problem we use PCA to bring the dimensions to a good balance between speed and accuracy. On the contrary, for other types of data, GHN is faster than KDtree.
By analyzing Table \ref{tab:times} data and Figure \ref{fig:times}, GHN is the best choice for real-time applications if you had to choose between GHN and KNN.

\begin{figure}[H]
	
	\centering
	
	\includegraphics[width=0.9\linewidth]{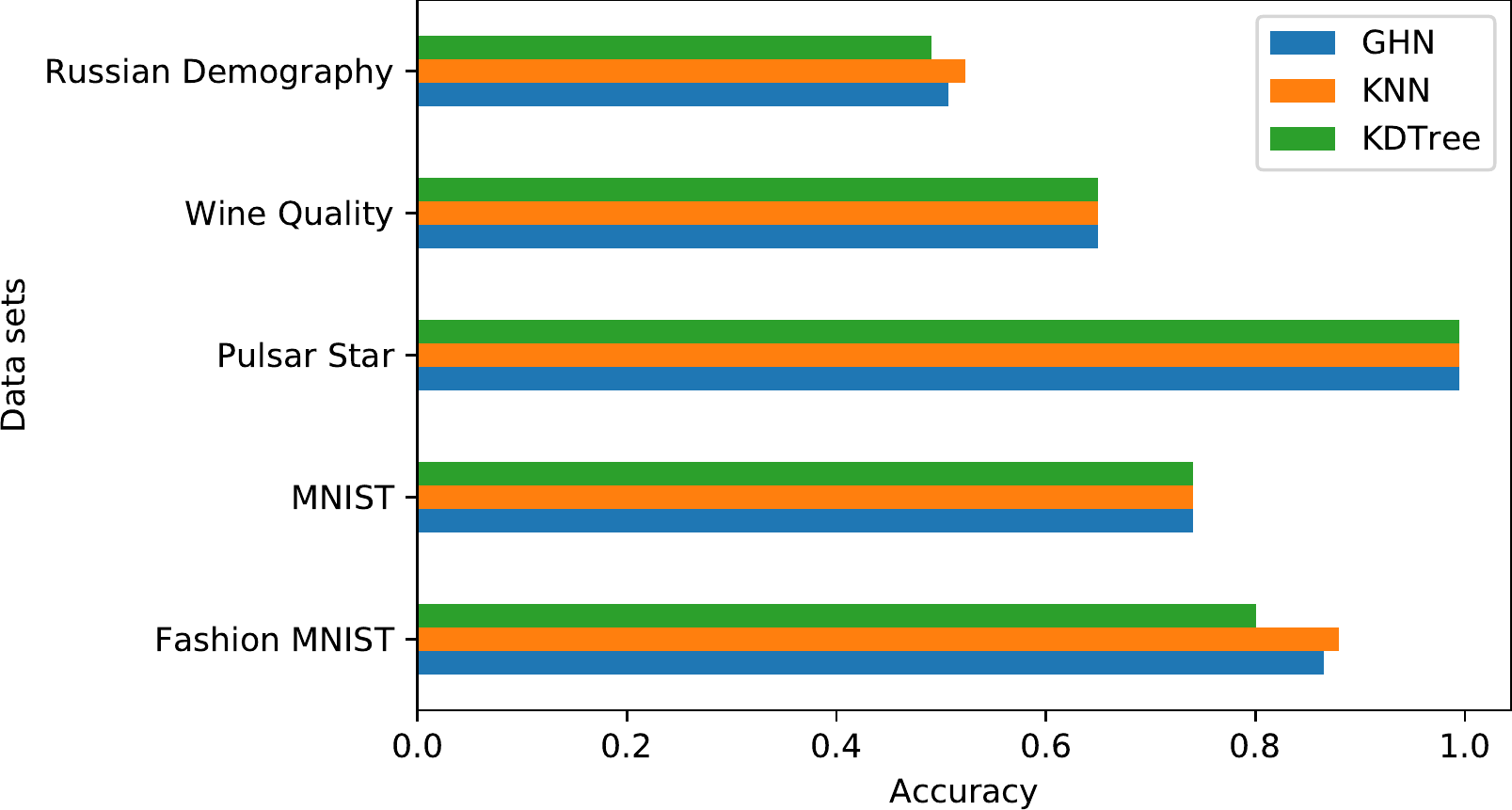}
	
	\caption{Comparison of accuracies on various datasets}
	
	\label{fig:acc}
	
\end{figure}

In Table \ref{tab:acc} and Figure \ref{fig:acc}, we compare accuracies of GHN,KDTree and KNN on all datasets.
\begin{table}[H]
	\centering
	\resizebox{0.7\linewidth}{!}{%
		\begin{tabular}{|l|l|l|l|} 
			\hline
			Datasets & GHN & KDTree & KNN \\ 
			\hline
			Fashion MNIST & 0.86 & 0.80 & 0.88 \\
			MNIST & 0.74 & 0.74 & 0.74~~ \\
			Pulsar Star & 0.99 & 0.99 & 0.99 \\
			Wine Quality & 0.65 & 0.65 & 0.65 \\
			Russian Demography & 0.50 & 0.49 & 0.52 \\
			\hline
		\end{tabular}
	}
	\caption{Accuracies on various datasets}
	\label{tab:acc}
\end{table}
These results show that the accuracies are almost identical except for some slight variations.
GHN offers better accuracy than KDtree on Russian Demography data and Fashion MNIST.
From the results of our experience, we confirmed that GHN can improve the time efficiency during predictions and it displays predictive accuracy as concurrent as that of KNN and KDTree. The main goal of GHN is to improve the prediction time by selecting the nearest neighbors directly in the neighborhood of our observation. Like KNN, GHN is sensitive to noise and is also subject to the curse of dimensionality. On the other hand, compared with proposed solutions till now that deal with the slowness of KNN, GHN is the only that is capable of achieving almost constant time performance when the observations located in a densely populated area of the data space.

\section{Conclusion}
\label{sec:Conclusion}
This paper proposes a new method for selecting the nearest neighbors to boost KNN prediction time which turns out to be very slow. The algorithm works in two steps: Cell sampling and Exploration. During the first step, a hash table is constructed to make the mapping between the data points of the training data and the cells of a virtual grid built on top of the data space. Finally, the second step which is done during predictions consists of using hashing to determine the cell to which our observation belongs. We search for the nearest neighbors starting from this cell and by visiting the cells around it. The experiment validated the performance of GHN which was faster than KNN during the evaluation on all our testing datasets. GHN is compatible for both regression and classification tasks with a prediction quality as good as that of KNN. GHN displays these good performances because, during the lookups, it looks for the nearest neighbor directly in the neighborhood of new observations, unlike KNN which needs to search throughout the entire dataset. But we emphasize that due to the effect of the curse of dimensionality which exponentially increases the number of cells from a layer to another during exploration, GHN is not recommended for data spaces with too many dimensions.

\appendices
\section{Proof of Equation \ref{equ:layer_cell_number}}

\label{AppendixA}

How many cells are on a given layer $l$?

Let's define by:

\begin{itemize}
	
	\item $a$ : cell measurements
	
	\item $l$ : the layer 
	
	\item $f$ : the number of features
	
\end{itemize}

With a look at the figure \ref{fig:gridknn}.c, we can deduce that the area from the central cell to a layer $l$ is :

\begin{eqnarray}
A(l)= [(2l+1)a]^f\nonumber
\end{eqnarray}

To only obtain the area covered by cells on layer $l$ only, we must, therefore, deduct from $A(l)$ the area $A(l-1)$:

\begin{eqnarray}
A(l) - A(l-1)= [(2l+1)a]^f - [(2(l-1)+1)a]^f\nonumber
\end{eqnarray}

Now we can compute the number of cells on layer l. For that we just have to divide $A (l) - A (l - 1)$ by the cell volume($a^f$):

\begin{eqnarray}
N_{cells} = \frac{A(l) - A(l-1)}{a^f}\nonumber\\
N_{cells} = \frac{[(2l+1)a]^f - [(2(l-1)+1)a]^f}{a^f}\nonumber\\
N_{cells} = \frac{[(2l+1)a]^f}{a^f} - \frac{[(2(l-1)+1)a]^f}{a^f}\nonumber\\
N_{cells} = (2l+1)^f - (2(l-1)+1)^f\nonumber\\
N_{cells} = (2l+1)^f - (2l-1)^f\nonumber
\end{eqnarray}

Hence Equation \ref{equ:layer_cell_number}.

\section{Proof of Equation \ref{equ:total_cell_number}}

\label{AppendixB}

In this Appendix, we want to find the number of cells from the first layer to a layer $l$.

We obtain the area from the first to layer $l$ by deducing from $A (l)$ defined in Appendix \ref{AppendixA} the volume of the central cell:

\begin{eqnarray}
A(l) - a^f = [(2l+1)a]^f - a^f\nonumber
\end{eqnarray}

Dividing this volume with the cell area gives us the total number of cells:

\begin{eqnarray}
S_{cells}(l) = \frac{A(l) - a^f}{a^f}\nonumber\\
S_{cells}(l) = \frac{[(2l+1)a]^f - a^f}{a^f}\nonumber\\
S_{cells}(l) = (2l+1)^f - 1\nonumber
\end{eqnarray}

Which gives Equation \ref{equ:total_cell_number}.

\section*{Acknowledgment}
I would like to thank the \href{https://wqu.org/programs/data-science}{WorldQuant University} which made this project possible by freely providing useful online learning resources on data science.
I also thank all reviewers for their insightful suggestions

\Urlmuskip=0mu plus 1mu\relax
\bibliographystyle{IEEEtran}

\end{document}